\documentclass{article}

    \usepackage[final]{neurips_2021}

\usepackage[utf8]{inputenc} 
\usepackage[T1]{fontenc}    
\usepackage{hyperref}       
\usepackage{url}            
\usepackage{booktabs}       
\usepackage{amsfonts}       
\usepackage{nicefrac}       
\usepackage{microtype}      
\usepackage{xcolor}         

\usepackage{wrapfig}

\usepackage{amssymb}
\usepackage{amsthm}
\usepackage{amsmath}
\usepackage{graphicx}
\usepackage{dsfont}
\usepackage{algorithm}
\usepackage{algorithmic}
\usepackage{hyperref}

\newcommand{\bsym}[1]{\ensuremath{\boldsymbol{#1}}}
\newcommand{\mat}[1]{\bsym{#1}}
\newcommand{\R}{\ensuremath{\mathbb{R}}}
\newcommand{\E}{\ensuremath{\mathbb{E}}}
\newcommand{\mdp}[1]{\ensuremath{\mathcal{#1}}}
\newcommand{\w}{\mat{\mathrm{w}}}

\newcommand{\argmax}{\ensuremath{\mathrm{argmax}}}

\newcommand{\david}[1]{}   
\newcommand{\chris}[1]{}  

\title{Successor Feature Neural Episodic Control}

\author{%
  David Emukpere, Xavier Alameda-Pineda, Chris Reinke\\
  RobotLearn  \\
  INRIA Grenoble, LJK, UGA \\
}

\begin{document}

\maketitle

\begin{abstract}
    A longstanding goal in reinforcement learning is to build intelligent agents that show fast learning and a flexible transfer of skills akin to humans and animals. This paper investigates the integration of two frameworks for tackling those goals: episodic control and successor features. Episodic control is a cognitively inspired approach relying on episodic memory, an instance-based memory model of an agent's experiences. Meanwhile, successor features and generalized policy improvement (SF\&GPI) is a meta and transfer learning framework allowing to learn policies for tasks that can be efficiently reused for later tasks which have a different reward function. Individually, these two techniques have shown impressive results in vastly improving sample efficiency and the elegant reuse of previously learned policies. Thus, we outline a combination of both approaches in a single reinforcement learning framework and empirically illustrate its benefits. \let\thefootnote\relax\footnote{Source code: \url{https://gitlab.inria.fr/robotlearn/sfnec}}
\end{abstract}

\section{Introduction}
The idea of building intelligent agents and systems that learn purely by interaction with their environment, known as reinforcement learning \citep{sutton_reinforcement_2018}, is an appealing approach to artificial intelligence with solid connections to neuroscience and psychology \citep{niv_reinforcement_2009, botvinick_deeprl_neuroscience_2020}. 
Reinforcement learning has generated significant interest both in the research community and in public awareness, especially in combination with deep learning \citep{lecun_deep_2015}, a paradigm known as deep reinforcement learning \citep{Arulkumaran2017ABS}.
It has given rise to impressive achievements in various contexts, including building champion game players \citep{silver_mastering_2016, vinyals_grandmaster_2019, openai_dota_2019, schrittwieser_mastering_2020}, and solving long-standing problems in biology \citep{jumper_highly_2021} to list a few.
However, human intelligence has defining characteristics lacking in state-of-the-art deep reinforcement learning systems. 

One important restriction of these systems is that they require significantly large amounts of data to learn~\citep{lake_building_2017, Tsividis2017HumanLI}, as they need a lot of (repeated) exposure to learn rules/concepts contained in data samples which manifests as \textit{slow} learning. 
In contrast, humans can learn quickly and efficiently, making use of little data. 
As pointed out by~\cite{botvinick_reinforcement_fast_and_slow_2019}, a source of slowness in deep reinforcement learning can be attributed to the requirement for incremental parameter adjustment in gradient-based optimization of deep neural networks.
A technique that has been proposed to tackle the data efficiency problem is Neural Episodic Control (NEC)~\citep{pmlr-v70-pritzel17a}.
Instead of gradually learning a representation of the solution, i.e.\ the expected future reward of an action in a certain situation, it stores observed experiences, i.e.\ the resulting rewards of an action, directly in a memory.
Encountering a similar situation again, the experiences in the memory are recalled to decide which action yields the best outcome. 
As a result, episodic control learns significantly faster than gradient-based techniques.


Another typical trait of human learning is the ability to seamlessly transfer knowledge across similar tasks leading to a faster learning process in new tasks.
This problem is typically tackled under the frameworks of meta~\citep{Hospedales2021MetaLearningIN} and transfer learning~\citep{Taylor2009TransferLF, zhu2020transfer}.
One such ability of humans is to reevaluate previously learned behaviors given a new task setting \citep{momennejad2017successor}.
For example, to reevalute all the possible ways you learned to drive home from work while maximize a new weighted  combination of using minimum time and having scenic views.
The framework of Successor Features and Generalized Policy Improvement (SF\&GPI) provides a mechanisms to replicate this human ability.
It decomposes the representation of learned behaviors in an environment dynamics part, i.e.\ what will happen when I do this behavior, and a reward part, i.e.\ how to evaluate this outcome.
Given a set of learned behaviors, i.e. their environment dynamics, and a new reward function the expected return for each behavior can be computed and the best behavior chosen. 

The central idea proposed in this paper is a framework combining NEC with SF\&GPI, which we call Successor Feature Neural Episodic Control (SFNEC). 
We hypothesize that this would provide advantages from both approaches by merging the learning speed conferred by episodic control with flexible transfer from SF\&GPI. 
We choose these two frameworks for the following reasons. 
First, episodic control has both a well-founded cognitive science inspiration~\citep{tulving_1972, Lengyel2007HippocampalCT} and displays impressive sample efficiency results in reinforcement learning tasks~\citep{Blundell2016ModelFreeEC, pmlr-v70-pritzel17a}. 
Likewise, for successor features, the elegance of the SF\&GPI framework and the connections of successor representation~\citep{dayan_improving_SR_1993, gershman_successor_2018} to neuroscience form the basis of our motivation. 
Additionally, a recent study~\citep{tomov_multi-task_2021} suggests that humans use a strategy similar to SF\&GPI for multi-task reinforcement learning.

To summarize, our main contributions are:
\begin{itemize}
    \item Introduction of SFNEC, a novel approach integrating sample-efficient learning using episodic control with meta learning using SF\&GPI
    \item Empirical validation of SFNEC by showing its advantage over baseline SF\&GPI, and NEC
\end{itemize}

\section{Background}

\subsection{Reinforcement Learning}
Reinforcement learning~\citep{sutton_reinforcement_2018} refers to a learning process where an agent attempts to maximize cumulative rewards it can obtain while interacting with its environment. 
Reinforcement learning problems are formalized as \textit{Markov Decision Processes} (MDPs)~\citep{puterman_markov_1994}. 
A MDP is a tuple ($\mdp{S}, \mdp{A}, p, \mdp{R}, \gamma$), where $\mdp{S}$ is the state space, $\mdp{A}$ is the action space, $p$ is the state transition probability distribution function $p(s_{t+1} | s_t, a_t)$ defining the probability of ending in state $s_{t+1} \in \mdp{S}$ after an agent takes action $a_t \in \mdp{A}$ in state $s_t \in \mdp{S}$ at the current time step $t$, and $\mdp{R}$ is the reward function associated with a transition $(s_t, a_t, s_{t+1})$. 
The goal of a reinforcement learning agent is to learn a \textit{policy} $\pi$, a mapping from states to actions, so as to maximize the expected sum of discounted rewards $G_t = \E^\pi[\sum_{j = 0}^{\infty} \gamma^{j}r_{t+j}]$, called the \textit{return}, where $r_{t+j}$ are the rewards received at each time step, and $\gamma \in [0, 1)$ is the discount factor used to determine how much weight is accorded to future rewards. 

\textit{Value function} based methods represent a large class of reinforcement learning algorithms based on classical \textit{dynamic programming}~\citep{bellman_dynamic_2010}. 
They learn a value function, here an \textit{action value function}, that can be recursively represented according to the Bellman equation: 
\begin{equation*}
    Q^\pi(s_t, a_t) = \E^\pi\left[\sum_{j = 0}^{\infty} \gamma^{j}r_{t+j}\right] = \E\left[r_t + \gamma Q^\pi(s_{t+1}, a_{t+1})\right]~.
\end{equation*}
The policy is then defined by maximizing the Q-function: $\pi(s) = \argmax_a Q^\pi(s,a)$.
A widely used method within this class of algorithms is $Q$-learning~\citep{watkins_q-learning_1992} trying to learn the optimal Q-function: $Q^*(s_t, a_t) = \E\big[r_t + \gamma~ \argmax_{a_{t+1}} Q^\pi(s_{t+1}, a_{t+1})\big]$. \david{should the $Q$ function inside the bracket also be $Q^*$?}

Classical $Q$-learning is restricted to small problems because it requires a table of all state-action pairs which becomes prohibitive or even unfeasible when attempting to scale to high dimensional state spaces. Thus, more recently, powerful function approximators such as deep neural networks are used which allow methods like $Q$-learning to scale to high dimensional state spaces, as exemplified in Deep Q-Network~\citep{Mnih2013DQN} and more recent variants.

\subsection{Episodic Control} 
Episodic memory~\citep{tulving_1972} is a model from the field of psychology, which refers to an autobiographical kind of memory about one's personal experiences. 
Likewise, episodic control~\citep{Lengyel2007HippocampalCT} implies the utilization of episodic memory for reinforcement learning by replaying stored action sequences from previous experiences.

\paragraph{Neural Episodic Control:}
Neural Episodic Control (NEC)~\citep{pmlr-v70-pritzel17a} is a computational model of episodic control. 
Central to NEC, is a memory structure called \textit{differentiable neural dictionary} (DND) which is a table $M_a$ of a pair of dynamically growing arrays of keys and values $(K_a, V_a)$ for each action $a \in \mdp{A}$. The keys here represent a learned representation of the agent state, while the values are $Q$-value estimates. To estimate the $Q$-value for a particular $(s, a)$ pair, a lookup is performed with the corresponding DND for action $a$ using a query key $h$, which is a lower-dimensional representation of $s$. The governing equation is: $Q(s, a) = \sum_{i} w_i v_i$, where $v_i$ corresponds to $Q$-values stored in $V_a$ and $w_i$ are weights corresponding to the result of a normalized kernel $k$ between the query key $h$ and keys $h_i$ in $K_a$ as follows $w_i =  k(h, h_i) / \sum_j k(h, h_j).$

Two techniques were employed to enable the scalability of this model. 
First, the number of elements involved in lookups was limited to the top $50$ nearest neighbours, efficiently found using a k-d tree. 
Second, the sizes of the DNDs were kept limited by removing the least recently used items. 
Furthermore, the values stored in memory were $N$-step Q-value estimates: \[ Q^{(N)}(s_t, a_t) = \E \left[ \sum^{N - 1}_{j = 0}\gamma^jr_{t + j} + \gamma^N\max_{a^\prime} Q(s_{t + N}, a^\prime) \right]~.\]
Updates to keys already found in the DND memory store while learning were done using $Q$-learning as $Q_i \leftarrow Q_i + \alpha(Q^{(N)}(s, a) - Q_i)$.

\subsection{Meta and Transfer Learning}

Meta \citep{li2018deep} and transfer learning \citep{Taylor2009TransferLF, Lazaric2012TransferIR,zhu2020transfer} refer to methods that allow knowledge learned from one or several tasks to be reused when faced with new tasks. 
In reinforcement learning, these tasks are defined by a set of MDPs $\mathcal{M}$.
To have a transfer between MDPs, some shared structure must exist between them. 
For example, consider an agent facing a set of navigation tasks in a sequence. 
Assuming the dynamics remain the same across these tasks, we can specify the desired behaviour by rewarding the agent to reach specific locations. 
Thus, specifying different tasks for an agent in this environment corresponds to different reward functions based on its location.  
In this paper, we are interested in this setting for transfer where successive tasks solved by an agent only differ in their reward functions.
A prominent method for this setting is the SF\&GPI framework~\citep{Barreto2017SuccessorFF}. 

\paragraph{Successor Features:}
Successor features (SF) are based on the idea of learning a value function representation that decouples rewards from environment dynamics~\citep{Barreto2017SuccessorFF}.
This is accomplished under the assumption that rewards are a linear combination of features $\mat{\phi}_t = \mat{\phi}(s_t, a_t, s_{t+1}) \in \R^n$ that depend on a state transition and a weight vector $\w \in \R^n$: $r_t = \mat{\phi}_t^\top \w$. 
Features describe the essential aspects of states for evaluating them with a reward function in a low-dimensional representation.
Each MDP in $\mathcal{M}$ has a different weight vector that defines its reward function, and the features are shared between all MDPs.
As a result, the $Q$-function rewrites as: 
\begin{equation*}
    Q^\pi(s_t, a_t) 
    ~=~ \E^\pi\left[ \sum_{i = t}^\infty \gamma^{i-t} r_i \right]
    ~=~ \E^\pi\left[ \sum_{i = t}^\infty \gamma^{i-t} \mat{\phi}_i^\top\w \right]
    ~=~ \mat{\psi}^\pi(s_t, a_t)^\top \w ~,    
\end{equation*} 
where $\mat{\psi}^\pi(s, a)$ are known as the successor features (SF) of ($s_t$, $a_t$) under policy $\pi$. 
Also, SF satisfy a Bellman equation: $\mat{\psi}(s_t, a_t) = \mat{\phi}_{t} + \gamma \E^\pi[\mat{\psi}^\pi(s_{t+1}, \pi(s_{t+1}))]$, and thus can be learned using conventional reinforcement learning methods.

\paragraph{Generalized Policy Improvement:}
Generalized policy improvement (GPI) is an operation to combine multiple policies, i.e.\ policies that were learned in previous tasks, to define a policy $\pi(s)$ for a new task: $\pi(s) \in \argmax_a \max_i Q^{\pi_i}(s, a)$. 
Using the SF decomposition $Q^{\pi_i}(s_t, a_t) = \mat{\psi}^{\pi_i}(s_t, a_t)^\top \w$, the GPI operator becomes $\pi(s) \in \argmax_a \max_i \mat{\psi}^{\pi_i}(s_t, a_t)^\top \w$.
As a result, the operator allows reevaluating old policies in a new task with reward weight vector $\w$ to chose the best action according to them.
Based on this, an algorithm for transfer using SF\&GPI was proposed in \citep{Barreto2017SuccessorFF} called SFQL.

\section{Method: The SFNEC Model}

\begin{figure}
    \centering
    \includegraphics[width=0.72\linewidth]{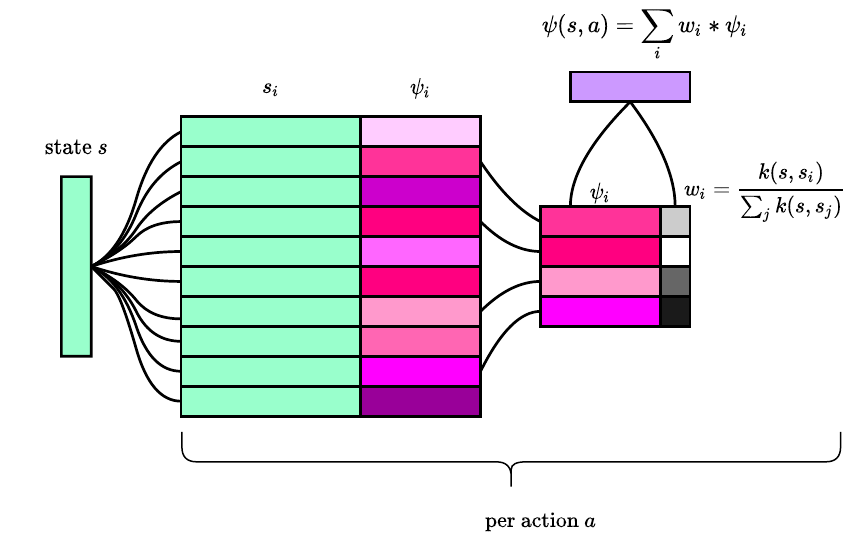}
    \caption{SFNEC architecture to store in an episodic manner $\mat{\psi}$-values.
    }
    \label{fig:sfnecmodel}
\end{figure}

Our proposed model, SFNEC (Fig.~\ref{fig:sfnecmodel}), extends NEC to learn successor features $\mat{\psi}(s,a) \in \R^n$ in place of scalar action-values. Like NEC, which learns $N$-step $Q$-values, SFNEC learns $N$-step $\mat{\psi}$-values:

\begin{equation}
    \label{eqn:nsteppsi}
    \mat{\psi}^{(N)}(s_t, a_t) = \E \left[ \sum_{j=0}^{N - 1} \gamma^j \mat{\phi}_{t + j} + \gamma^N \max_{a^\prime}\mat{\psi}(s_{t + N}, a^\prime) \right] ~.
\end{equation}

To perform a lookup using the SFNEC model, we use Equation~\ref{eqn:psilookup}:

\begin{equation}
    \label{eqn:psilookup}
    \mat{\psi}(s_t, a) = \sum_{i} \frac{k(s_t, s_i)}{\sum_{j} k(s_t, s_j)} * \mat{\psi}_i ~,  
\end{equation}

where $\mat{\psi}_i$ corresponds to a previously stored $\mat{\psi}_i$-value for a  state $s_i$ in memory, and $k$ is the kernel used to compute a similarity score between the query state $s_t$ and states in memory $s_i$. 
We note that we directly used the state vector $s_t$ as keys in memory for our experiments.
In general, the NEC architecture allows training an embedding network to learn a lower-dimensional state embedding to be used as keys in memory. Similar to NEC, we limit the elements in memory used during lookups to the top nearest neighbours, e.g., $50$. Likewise, we also used the inverse distance kernel used in \citep{pmlr-v70-pritzel17a}: $k(s, s_i) = \frac{1}{\|s-s_i\|_2^2 + \delta}$. 

During training, $\mat{\psi}$-values are updated after observing $N$ transitions. When the $\mat{\psi}$-value for a state action pair $(s, a)$ does not exist previously in memory, $N$-step estimates computed using Equation~\ref{eqn:nsteppsi} are inserted in the corresponding DND for action $a$. On the other hand, $\mat{\psi}$-values already in memory are updated using: $\mat{\psi}_i\leftarrow \mat{\psi}_i + \alpha(\mat{\psi}^{(N)}(s, a) - \mat{\psi}_i)$. 


We defer for details of the algorithm and the training procedure to Appendix~\ref{app:agents}.

\newpage

\section{Experiments}
\label{sec:experiments}

\begin{wrapfigure}{r}{0.5\textwidth}

    \vspace{-4mm}
    \centering
    \includegraphics[width=0.65\linewidth]{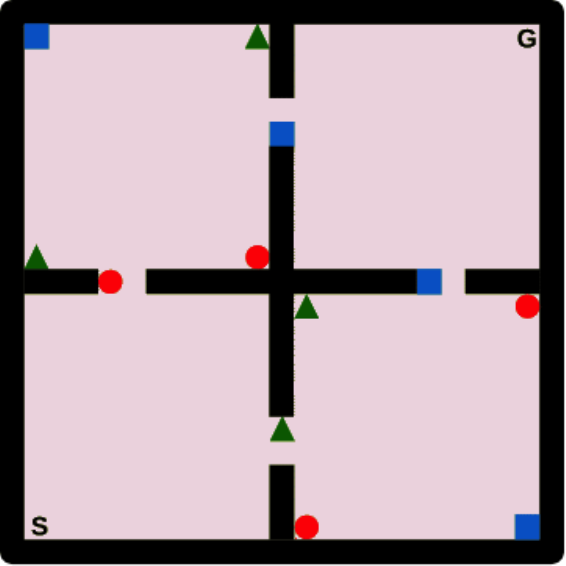}
    \caption{
    \chris{You need your own figure for the final version to avoid copyright issues!}
    2-D object collection environment proposed in \citep{Barreto2017SuccessorFF}. 
    }
    \label{fig:object_collection_task}
\end{wrapfigure}

\chris{Describe the connections between tasks, rewards, weights and features.}

We evaluated SFNEC in the two-dimensional object collection environment (Fig.~\ref{fig:object_collection_task}) proposed by \cite{Barreto2017SuccessorFF}. 
The environment consists of four rooms with a start location in the bottom left denoted 'S', and a goal location in the top right denoted 'G'. 
Multiple objects exist within the rooms belonging to three classes shown as a circle, square, and triangle. 
The goal is to navigate from the start position to the goal position while picking up objects to maximize the cumulative rewards. 
Objects once picked up disappear and reappear at the beginning of a new episode. 
To utilize the SF\&GPI framework as defined in~\citep{Barreto2017SuccessorFF}, it is necessary to have a linear decomposition of rewards into features and weights as $r_t = \mat{\phi}^{\top} \w$. 
The features describe the picked up object classes and if the goal state is reached using a binary representation: $\mat{\phi} \in \{0, 1\}^4$. 
The first three feature components represent if an object belonging to one of the three classes has been picked, while
the last component represents if the goal state is reached. 
Thus, rewards associated with each transition can be expressed as a dot product between these features and weight vectors $\w \in \R^4$ that contain the reward associated with picking up each object class and reaching the goal. Different tasks in the environment are then defined by setting a weight vector $\w$.

To demonstrate good performance, an agent faces a series of tasks, each being a different instantiation of $\w$ with the aim of maximizing the sum of rewards accumulated by the agent. In general, we follow the same setup for this environment as in \citep{Barreto2017SuccessorFF} with further details given in Appendix~\ref{app:experiments}. 
We compare SFNEC with SFQL, NEC, and a version of SFNEC without GPI.

\paragraph{Results:}
\label{subsec:four_room_results}

\begin{figure}[htbp]
    \centering
    \includegraphics[width=\textwidth]{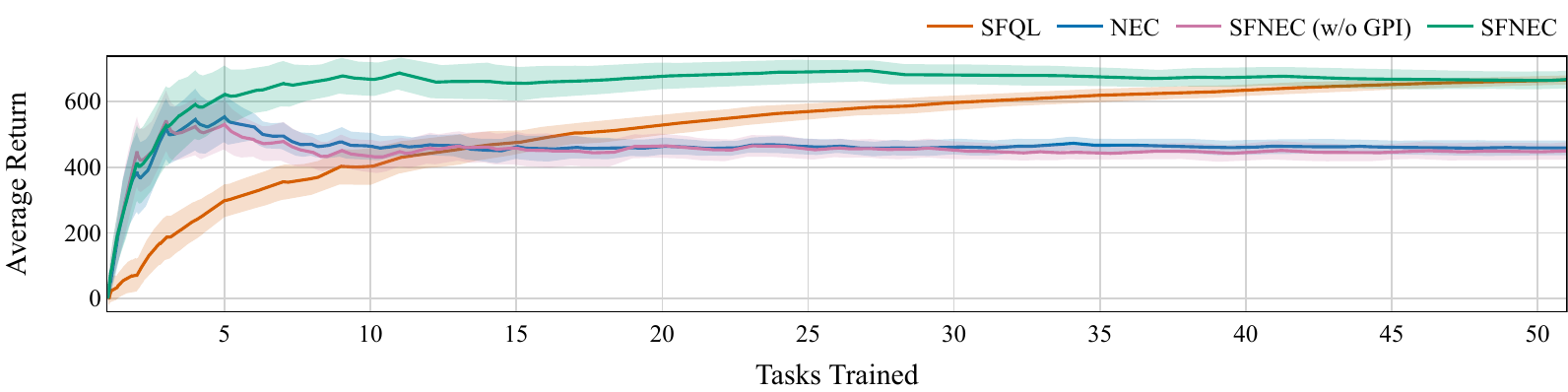}
    \caption{
    SFNEC has a higher learning speed than other methods. 
    Depicted is the mean of the average return per task over 10 runs with the standard error shown as the shaded regions.}
    \label{fig:four_room_results}
\end{figure}

We compared the average return per task over $10$ runs of each algorithm (Fig.~\ref{fig:four_room_results}). 
SFNEC with GPI performs best, outperforming NEC and SFQL agents. 
We expect this as SFNEC combines learning speed from episodic control on each task with the strong transfer conferred by SF\&GPI. 

Furthermore, we note that NEC and SFNEC without GPI show a significantly improved learning speed over the SFQL baseline during the first $10$ tasks due to their episodic memory even without having a transfer mechanism.

\section{Discussion}
\label{chap:discussion}

\paragraph{Complementary benefits of episodic control and SF\&GPI:} 
Like SFQL, SFNEC learns $\mat{\psi}$-values and utilizes GPI. 
However, SFNEC uses episodic memory, which means the main advantage we expect would be improved learning speed compared to SFQL. This is confirmed in the results (Fig.~\ref{fig:four_room_results}) as SFNEC rises in performance much faster than SFQL, even though they reach similar performance levels in the long run due to both algorithms utilizing GPI for transfer. 
On the other hand, NEC and SFNEC without GPI also employ episodic control but differ from SFNEC by not utilizing GPI but attempting to learn each task from scratch rapidly. More specifically, NEC directly learns $Q$-values, while SFNEC without GPI learns $\mat{\psi}$-values. 
We observe that SFNEC without GPI and NEC perform similarly and can demonstrate reasonably strong performance in learning all tasks individually due to their usage of episodic control. However, they are not able to match the long-run performance of SFNEC. 
Putting both comparisons together, we deduce that the combination of episodic control with transfer using the SF\&GPI framework in SFNEC brings together rapid learning and transfer. However, as can be seen towards the right end of Fig.~\ref{fig:four_room_results}, SFQL would most probably overtake SFNEC in subsequent phases. 
This is similar to the observation reported in~\cite{pmlr-v70-pritzel17a} where parametric methods like DQN outperform NEC in the long run. 
We leave an investigation of integrating methods proposed to tackle this into SFNEC for subsequent work.

\paragraph{Learning a lower-dimensional state embedding:}
We conducted further preliminary experiments with SFNEC that attempted to learn a lower-dimensional state embedding for the keys of the DND, like the original setup in NEC. However, this did not yield good results yet. We hypothesize that this might be due to higher approximation errors introduced when learning an embedding. 
With the method we used in this paper, where we did not learn an embedding, we have the advantage that the keys are stable as they come directly from environment observations. 
Thus, learning is easier for the agent as it is not burdened with a representation learning stage. Logically, we can expect that if the agent cannot learn a good embedding to produce keys in the DND for a particular task, then reusing this on a new task can lead to erroneous predictions of the $\mat{\psi}$-values. 
Essentially, this would mean the approximation error could be high for the policies involved in the GPI procedure because of inaccurately learned embeddings, which could result in poor performance.

\section{Related work}
\label{chap:related}

\paragraph{Episodic Control:}
Several improvements and extensions to NEC have been proposed. 
In \citep{lin_episodic_2018} Episodic Memory Deep Q Network was proposed, an architecture that augments DQN with an episodic memory-based estimate. 
They found that combining this with a TD estimate improved sample efficiency and long-term performance over DQN and NEC. 
\cite{sarrico_sample-efficient_2019} investigated adding principled exploration to NEC by combining episodic control with maximum entropy mellowmax policy. 
\citep{agostinelli_memory-efficient_2019} on the other hand, proposed using dynamic online k-means to improve the memory efficiency of NEC. 
Likewise,~\citep{Zhu2020EpisodicRL} proposed to further optimize the usage of the contents of the episodic memory store by considering the relationship between contents of episodic memory. Finally,~\citep{Hu2021GeneralizableEM} recently introduced \textit{Generalizable Episodic Memory} which extends the applicability of episodic control to continuous action domains.
These extensions are parallel developments to SFNEC and could be integrated with it.

\paragraph{Successor Features:}
A few extensions to successor features exist. 
A relaxation of the condition that reward functions be expressed as a linear decomposition, as well as a demonstration of how to combine deep neural networks with SF\&GPI was introduced in \citep{Barreto2019SF}. 
Another direction aims to learn appropriate features from data such as by optimally reconstruct rewards \citep{Barreto2017SuccessorFF}, using the concept of mutual information \citep{hansen2019fast}, or the grouping of temporal similar states \citep{madjiheurem2019state2vec}.
A further direction is the generalization of the $\psi$-function over policies \citep{borsa2018universal} analogous to universal value function approximation \citep{schaul2015universal}. 
Similar approaches use successor maps \citep{madarasz2019better}, goal-conditioned policies \citep{ma2020universal}, or successor feature sets \citep{brantley2021successor}.
However, none of these extensions studied the usage of SF in combination with episodic memory.

\section{Conclusion}
\label{chap:conclusion}
We introduced SFNEC, and showed its viability as a framework that combines rapid learning and transfer. However, a few problems would need to be addressed to obtain a robust practical implementation. An example would be investigating methods for reducing memory requirements as this is a real-world constraint. Similarly, deciding when to learn tasks or automatically detect task switches is an area to tackle.
Some suggestions for tackling such problems have been pointed out in \citep{Barreto2017SuccessorFF}, and we believe it would be fruitful work to investigate applying SFNEC on real-world tasks with an evaluation of different techniques to handle these various challenges.

\section*{Acknowledgment}

This research was supported by the H2020 SPRING project funded by the European Commission under the Horizon 2020 framework programme for Research and Innovation (H2020-ICT-2019-2, GA \#871245) as well as by the ML3RI project funded by French Research Agency under the Young Researchers programme \#ANR-19-CE33-0008-01.

\bibliography{references.bib}

\newpage
\appendix


\section{Supplemental Experiments}

We now discuss a few supplemental experiments run to understand our model.

\begin{figure}[H]
    \centering
    \includegraphics[width=0.65\linewidth, height=4cm]{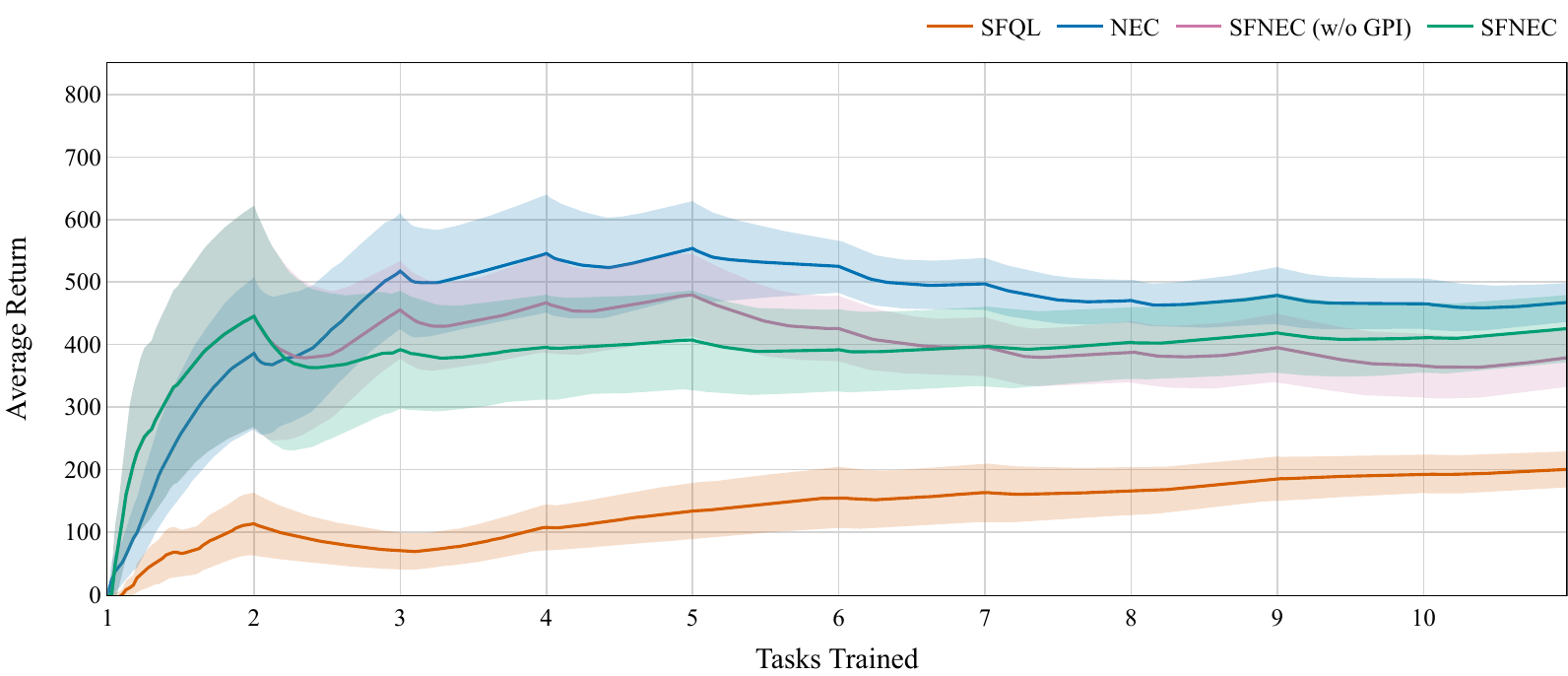}
    \caption{Average return of SFQL, NEC, and SFNEC agents on the four-room environment while learning $\w$. Performance drops for SFQL and SFNEC agents that depend on $\w$ compared to Figure~\ref{fig:four_room_results}. NEC performance remains the same and performs best in this setting. Averages are taken over 10 runs with the standard error shown as the shaded region around solid lines.}
    \label{fig:w_not_given}
\end{figure}

\paragraph{What is the effect of learning $\w$?}
Here, we run experiments where the reward weight vector $\w$ is not provided to agents; rather, it is approximated while interacting with the environment for algorithms that need $\w$ i.e., SFQL and SFNEC. As shown in Figure~\ref{fig:w_not_given}, we noticed a reduction in the performance across all agents that rely on $\w$.
 
We posit that approximation errors from estimating $\w$ and the successor features $\mat{\psi}$ might lead to this reduced performance for agents using successor features, especially as the best performing agent in this setup is the NEC agent that does not rely on $\w$ nor successor features. Nonetheless, there are many application domains where the reward function given by $\w$ would be known, and the SFNEC model with GPI would perform best in this case.

\paragraph{What is the effect of varying memory capacity?}
We know an essential consideration in a real-life implementation of our proposed SFNEC model and algorithm is how well it will scale with many tasks. However, the scheme scales linearly, and this might be too expensive if one keeps a large DND memory per action per task. Thus, we were interested in observing the degradation of the agent's performance with the DND capacity.

\begin{figure}[H]
    \centering
    \includegraphics[width=0.65\linewidth, height=4cm]{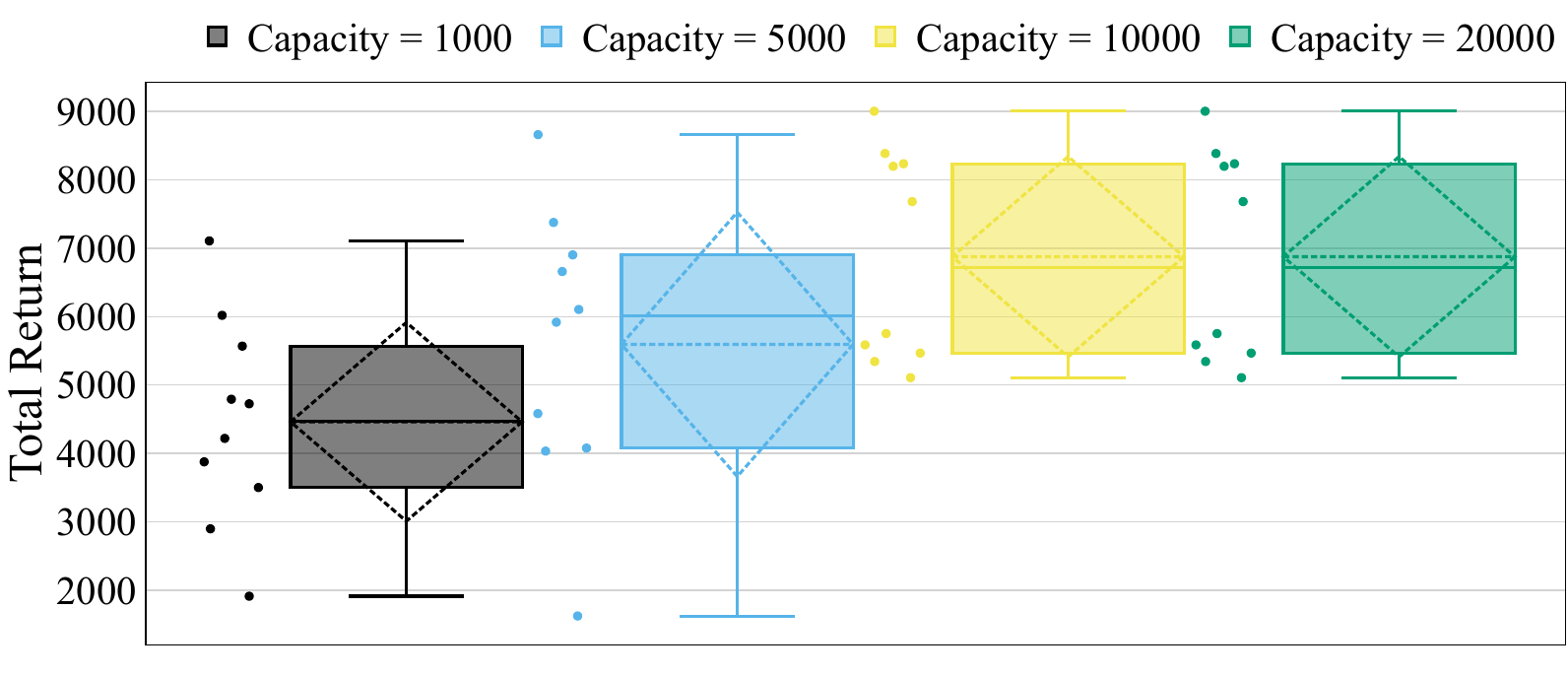}  
    \caption{Total return over 10  runs of SFNEC agents on the four-room environment when varying the DND memory capacity.  Beyond 10000, further increase in capacity does not lead to improved performance.}
    \label{fig:dnd_capacity}
\end{figure}

As shown in Figure~\ref{fig:dnd_capacity}, there is a point beyond which increasing capacity does not lead to improvement in performance. Practically, this means we might be able to use the minimum possible capacity for memory that guarantees good performance, knowing that a larger capacity would not result in performance gains. Also, the gradual degradation in performance shows that the method can be flexibly tuned to achieve the desired tradeoff between performance and memory requirements for a particular application. 

\begin{figure}[H]
    \centering
    \includegraphics[width=0.65\linewidth, height=4cm]{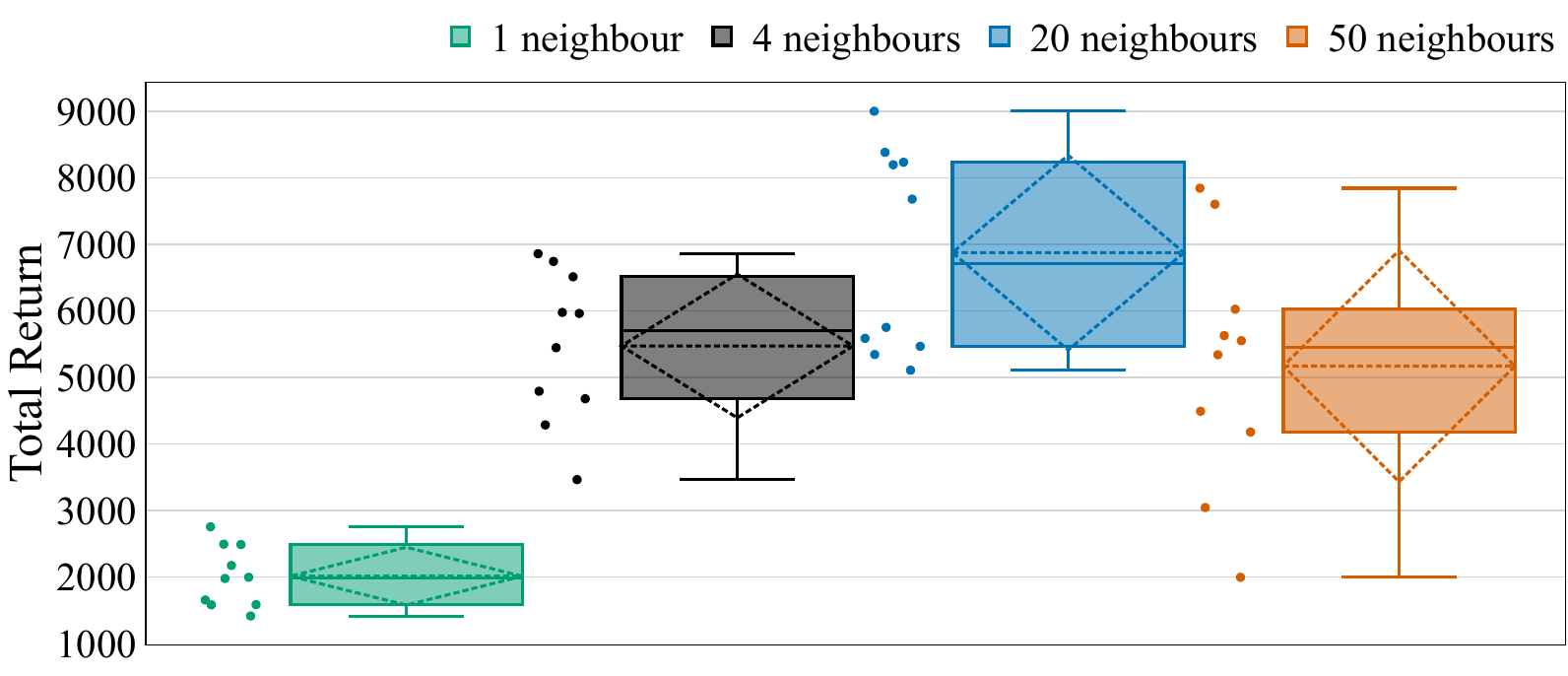}  
    \caption{\small{Total Return over 10 random runs of SFNEC agents on the four-room environment when varying number of neighbours used for value estimation. The plot depicts more that the performance follows a 'U-curve'.}}
    \label{fig:dnd_neighbours}
\end{figure}

\paragraph{What is the effect of varying number of neighbours?}
A critical parameter when performing value estimation using our model is the number of neighbours used. This induces a sort of bias-variance tradeoff, and we experimented with different settings shown in Figure~\ref{fig:dnd_neighbours}.

Our results indicate that varying the number of neighbours should follow a 'U-curve' for a particular application, reducing performance when going towards either extreme of estimating with a single neighbour or the entire memory. This is expected because,  on one end, we will have a method that obtains estimates that overfits to its nearest neighbour (i.e. high variance). On the other end, we have a method that tries to fit its estimate over the entire dataset provided by the memory module (i.e., high bias).

\section{Agents}
\label{app:agents}
In this section, we give a few more details on the agents in this paper. Source code is available at \url{https://gitlab.inria.fr/robotlearn/sfnec}.

\paragraph{SFQL agent:}
As our baseline agent for transfer learning, we used an implementation of SF\&GPI according to the pseudocode provided in~\citep{Barreto2017SuccessorFF} which we call SFQL agent.

\paragraph{NEC agent:}
We implement an agent following the Neural Episodic Control architecture and algorithm outlined in \citep{pmlr-v70-pritzel17a} as the baseline for learning speed on a single task. We relied on a publicly available implementation provided by Kai Arulkumaran on GitHub\footnote{\url{https://github.com/Kaixhin/EC}} as an inspiration for the implementation of our NEC agent. For our purposes, we modified our implementation to allow memory updates to occur either immediately after a horizon of $N$ steps or batching updates at the end of episodes. This is different from the original description and the reference implementation, which suggest batching memory updates at the end of episodes. Our primary motivation for this modification is that we would like our method to be applicable even in learning scenarios that do not divide into episodes, i.e., continuing tasks. 

Estimating action values with NEC involves combining the values for previously-stored keys in memory similar to a query key representing the agent's state. Thus, it is necessary to perform an efficient similarity search.  In~\citep{pmlr-v70-pritzel17a} it was suggested to enable this efficiency by performing approximate searches using a k-d tree~\citep{Bentley1975}. Contrary to this, to keep our initial implementation simple while laying out the general idea of our framework, we utilize brute-force searches for all agents that need to perform a similarity search in stored memory(NEC and SFNEC). 

For this, we used the Facebook AI similarity search library\footnote{\url{https://github.com/facebookresearch/faiss}}~\citep{JDH17}. We chose to use this library because it is open-source, simple to use, and well-optimized for brute-force and approximate nearest neighbour searches.

\paragraph{SFNEC agent:}
The algorithm for SFNEC is given in Algorithm~\ref{alg:sfnecalgo}. We note that we allow for both cases when reward descriptions $\w_i$ for each task are provided or not with the boolean condition learn\_w. Additionally, we note that we can use the algorithm without GPI by simply making $j$ equal $i$ in line~\ref{it:gpi_action_selection} similar to the SFQL algorithm described in \citep{Barreto2017SuccessorFF}. 

Finally, we highlight that we also update policies used for GPI action selection in lines~\ref{it:update_prev_start}-\ref{it:update_prev_end} as done in the SFQL algorithm. The essence of this update is to continually refine the successor features of policies that remain pertinent for GPI action selection in line~\ref{it:gpi_action_selection}. A crucial difference for updating these policies is that we cannot obtain the features $\mat{\phi}$ needed to compute $N$-step $\mat{\psi}$ estimates as the agent is acting according to a different policy at this update point. Thus, we are constrained to utilizing a single-step \textit{off-policy} update. 

\begin{algorithm}
    \caption{Successor Feature Neural Episodic Control (SFNEC)
		\label{alg:sfnecalgo}}
	\begin{algorithmic}[1]
	\REQUIRE
	\begin{tabular}{cl}
    $\alpha_w$ & learning rate for $\w$ \\
    learn\_w & condition to indicate if to learn $\w$ for each task \\
    $\mat{\phi}$ & features of state transitions \\
    $\w_i$ & optionally given for each task $i$ \\
    $N$ & n-step horizon for $\psi^{(N)}$ estimates \\
    $\mathcal{D}_i$ & replay buffer of ($h$, $a$, $\psi^{(N)}$) tuples for each task $i$ \\
    $M_{ai}$ & a DND for each action $a$ per task $i$ \\
    num\_tasks & number of tasks to be learned \\ ~ \\
	\end{tabular}
    \FOR{$i = 1, \dots, \text{num\_tasks}$}
        \IF{learn\_w}
        \STATE Initialize $\w_i$ with small random values
        \ENDIF
    	\FOR{each episode}
            \FOR{$t = 1, \dots, T$}
                \STATE Get observation $s_t$ from the environment and its embedding $h_t$
               \STATE  $j \leftarrow \argmax_{k \in \{1, \dots, i\}}
                \max_{b} \mat{\psi}_k(s_t,b)^{\top}\w_i$
                \label{it:gpi_action_selection}
                 \\ \hfill \COMMENT{j is the index of policy selected according to GPI}
                
                \IF{$rand[0, 1) < \epsilon$}
                
                \STATE $a_t \leftarrow $ select an action uniformly at random
                \ELSE
                \STATE $a_t \leftarrow \argmax_b \mat{\psi}_{j}(s_t, b)^{\top}\w_i$
                \ENDIF
                \hfill \COMMENT{$\epsilon$-greedy action selection} 
               
                \STATE Take action $a_t$ and observe reward $r_t$, and observation $s_{t+1}$ \\ ~ \\
                \IF{learn\_w}
                \STATE $\w_i \leftarrow \w_i + \alpha_w[r - \mat{\phi}(s_t, a_t, s_{t+1})^{\top}\w_i]$ \hfill \COMMENT{learn $\w$ for task $i$}
                \ENDIF
                \STATE Compute $\mat{\psi}^{(N)(s_t,a_t)}$ using eqn.~\ref{eqn:nsteppsi} \label{it:compute_psi}
                \STATE Append $(h_t,\mat{\psi}^{(N)}(s_t,a_t))$ to $M_{a_ti}$ \label{it:update_mem_start}
                \STATE Append $(s_t, a_t, \mat{\psi}^{(N)}(s_t,a_t))$ to $\mathcal{D}_i$ \label{it:update_mem_end}
                \STATE Train on a random minibatch from $\mathcal{D}_i$
                \label{it:minibatch} \\ ~ \\
                
                \IF{$j \neq i$} \label{it:update_prev_start}
                \STATE $a' = \argmax_b \mat{\psi}_j(s_{t+1}, b)^{\top}\w_j$
                \STATE update $\mat{\psi}_j(s_t, a_t)$ using the one-step TD target: $\mat{\phi}(s_t, a_t, s_{t+1}) + \gamma\mat{\psi}_j(s_{t+1}, a')$
                \ENDIF \label{it:update_prev_end}
    	    \ENDFOR
    	\ENDFOR
    \ENDFOR
	\end{algorithmic}
\end{algorithm}

\section{Experimental details}
\label{app:experiments}
As mentioned in the main paper, we follow the same setup as described in~\citep{Barreto2017SuccessorFF} with essential details recapitulated below. There are twelve objects in the environment and three object classes (four objects per class). The rewards associated with each class changes after $20, 000$ transitions, and they are sampled uniformly at random from $[-1, 1]$ while reaching the goal always gave a reward of $+1$. The agent's observations provided from the environment consists of two parts. The first part is the activations of the agent's $(x, y)$ position on a $10 \times 10$ grid of radial basis functions over the entire four rooms. The second part consists of object detectors indicating the presence or absence of objects in the environment. For our experiments, we provide both the state features $\mat{\phi}(s_t, a_t, s_{t + 1})$ and reward weight vector $\w$. The state features are boolean vectors containing elements indicating whether the agent is over an object present in the environment or over the goal position. The reward weight vector contains rewards associated with picking up an object and reaching the goal position.  We note that providing both elements to the agent means the reward function is fully specified to the agent according to the decomposition: $r(s_t, a_t, s_{t + 1}) = \mat{\phi}(s_t, a_t, s_{t + 1})^\top\w$. Nonetheless, it is possible to approximate these quantities as shown in~\citep{Barreto2017SuccessorFF}. We refer the reader to Appendix B in~\citep{Barreto2017SuccessorFF} for further details on this environment. 

\paragraph{Setup:}
We use $50$ tasks for our experiments. For obtaining our hyperparameters, we carried out a grid search for the NEC and SFNEC agents. At the same time, we relied on reported values for SFQL agent parameters from \citep{Barreto2017SuccessorFF}. We now report values used for our search and the final values chosen for our experiments. We tested values in $\{0.05, 0.15\}$ for $\epsilon$ used for $\epsilon$-greedy exploration. For the learning rate used in network optimization: $\{0.01, 0.05, 0.1\}$, for number of neighbours used in estimating $\psi$-values: $\{1, 4, 10, 20, 50\}$. Furthermore, we limited the DND capacity to $10, 000$ most recent entries. For the fast learning rate used to update re-encountered keys in the DND, we tried values in $\{0.1, 0.3, 0.5\}$, and for horizon length $N$, we used the set: $\{8, 16, 32\}$. We show the best configurations found for each agent in Table~\ref{table:configurations}.

For our NEC and SFNEC agents, we obtain the keys used as the compact state representation in memory by directly using the observation from the environment. Additionally, the training method we use for NEC and SFNEC keeps a single sample in the replay buffer of these agents. Training proceeds by using each sample sequentially as they are collected in the environment. We did this to allow a more direct comparison to SFQL, which uses a "true" stochastic gradient descent method for its optimization.

\begin{table}[h!]
\centering
\begin{tabular}{lccccc}
\toprule
Agent & $\epsilon$ & Network learning rate & Neighbours & DND learning rate & $N$\\  
 \midrule
 SFQL & 0.15 & 0.01 & - & - & - \\ 
 NEC & 0.15 & 0.01 & 20 & 0.1 & 8 \\
 SFNEC & 0.15 & 0.05 & 20 & 0.1 & 8 \\
 \bottomrule
\end{tabular}
\caption{Best parameter configurations found for agents in the four-room environment}
\label{table:configurations}
\end{table}

\end{document}